\documentclass[runningheads]{llncs}
\usepackage{graphicx}
\usepackage{comment}
\usepackage{multirow}
\usepackage{amsmath,amssymb} 
\usepackage{color}

\begin{document}
\pagestyle{headings}
\mainmatter
\def\ECCVSubNumber{3948}  

\title{EagleEye: Fast Sub-net Evaluation for Efficient Neural Network Pruning} 

\titlerunning{EagleEye}
%
\author{Bailin Li\inst{1} \and
Bowen Wu\inst{2} \and
Jiang Su\inst{1}
\and
Guangrun Wang\inst{2}
\and Liang Lin\inst{1,2}}
\authorrunning{Li et al.}
%

\institute{Dark Matter AI Inc. \and Sun Yat-sen University \\
\email{bl-zorro@163.com}, \email{\{wubw6,wanggrun\}@mail2.sysu.edu.cn}, \email{sujiang@dm-ai.cn},
\email{linliang@ieee.org}}

\maketitle

\begin{abstract}
Finding out the computational redundant part of a trained Deep Neural Network (DNN) is the key question that pruning algorithms target on. Many algorithms try to predict model performance of the pruned sub-nets by introducing various evaluation methods. But they are either inaccurate or very complicated for general application. In this work, we present a pruning method called EagleEye, in which a simple yet efficient evaluation component based on adaptive batch normalization is applied to unveil a strong correlation between different pruned DNN structures and their final settled accuracy. This strong correlation allows us to fast spot the pruned candidates with highest potential accuracy without actually fine-tuning them. This module is also general to plug-in and improve some existing pruning algorithms. EagleEye achieves better pruning performance than all of the studied pruning algorithms in our experiments. Concretely, to prune MobileNet V1 and ResNet-50, EagleEye outperforms all compared methods by up to 3.8\%. Even in the more challenging experiments of pruning the compact model of MobileNet V1, EagleEye achieves the highest accuracy of 70.9\% with an overall 50\% operations (FLOPs) pruned. All accuracy results are Top-1 ImageNet classification accuracy. Source code and models are accessible to open-source community.\footnote{https://github.com/anonymous47823493/EagleEye}
\end{abstract}

\section{Introduction}
\label{sec:intro}
Deep Neural Network (DNN) pruning aims to reduce computational redundancy from a full model with an allowed accuracy range. Pruned models usually result in a smaller energy or hardware resource budget and, therefore, are especially meaningful to the deployment to power-efficient front-end systems. However, how to trim off the parts of a network that make little contribution to the model accuracy is no trivial question.

DNN pruning can be considered as a searching problem. The searching space consists of all legitimate pruned networks, which are referred as sub-nets or pruning candidates. In such space, how to obtain the sub-net with highest accuracy with reasonably small searching efforts is the core of a pruning task.

\begin{figure}[h]
    \centering
       \includegraphics[width=0.63\linewidth]{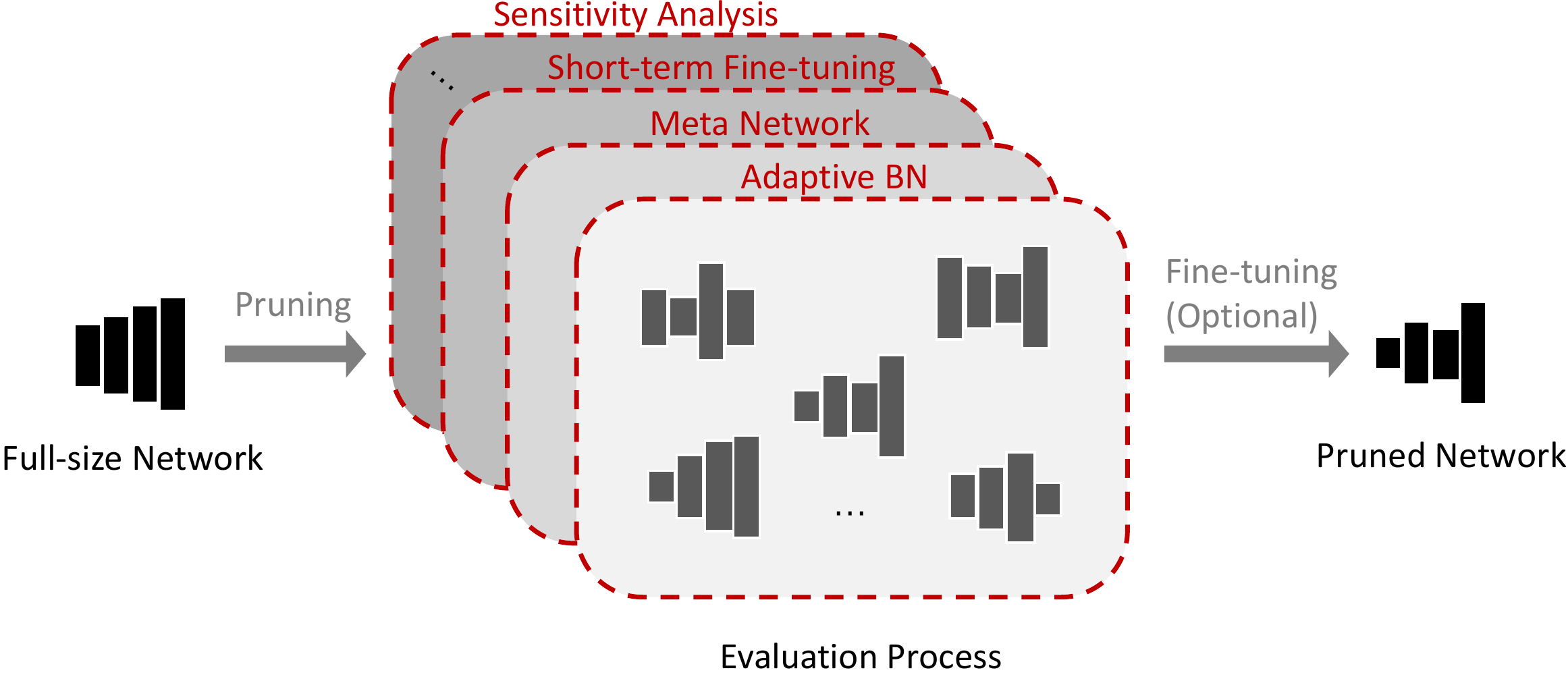}
       \caption{A generalized pipeline for pruning tasks. The evaluation process unveils the potential of different pruning strategies and picks the one that most likely to deliver high accuracy after convergence.}
       \label{fig:cover}
\end{figure}

Particularly, an evaluation process can be commonly found in existing pruning pipelines. Such process aims to unveil the potential of sub-nets so that best pruning candidate can be selected to deliver the final pruning strategy. A visual illustration of this generalization is shown in Figure~\ref{fig:cover}. More details about the existing evaluation methods will be discussed throughout this work. An advantage of using an evaluation module is fast decision-making because training all sub-nets, in a large searching space, to convergence for comparison can be very time-consuming and hence impractical.

However, we found that the evaluation methods in existing works are sub-optimal. Concretely, they are either inaccurate or complicated.

By saying “inaccurate”, it means the winner sub-nets from the evaluation process do not necessarily deliver high accuracy when they converge~\cite{fp,amc,autocompress}. This will be quantitatively proved in Section~\ref{sec:corr} as a correlation problem measured by several commonly used correlation coefficients. To our knowledge, we are the first to introduce correlation-based analysis for sub-net selection in pruning task. Moreover, we demonstrate that the reason such evaluation is inaccurate is the use of sub-optimal statistical values for Batch Normalization (BN) layers~\cite{bn}. In this work, we use a so-called “adaptive BN” technique to fix the issue and effectively reach a higher correlation for our proposed evaluation process.

By saying “complicated”, it points to the fact that the evaluation process in some works rely on tricky or computationally intensive components such as a reinforcement learning agent~\cite{amc}, auxiliary network training ~\cite{thinet}, knowledge distillation~\cite{cp}, and so on. These methods require careful hyper-parameter tuning or extra training efforts on the auxiliary models. These requirements make it potentially difficult to repeat the results and these pruning methods can be time-consuming due to their high algorithmic complexity.

Above-mentioned issues in current works motivate us to propose a better pruning algorithm that equips with a faster and more accurate evaluation process, which eventually helps to provide the state-of-the-art pruning performance. The main novelty of the proposed EagleEye pruning algorithm is described as below:

\begin{itemize}
    \item We point out the reason that a so-called “vanilla” evaluation step (explained in Section \ref{sec:motivation}) widely found in many existing pruning methods leads to poor pruning results. To quantitatively demonstrate the issue, we are the first to introduce a correlation analysis to the domain of pruning algorithm. 
    \item We adopt the technique of adaptive batch normalization for pruning purposes in this work to address the issue in the “vanilla” evaluation step. It is one of the modules in our proposed pruning algorithm called EagleEye. Our proposed algorithm can effectively estimate the converged accuracy for any pruned model in the time of only a few iterations of inference. It is also general enough to plug-in and improve some existing methods for performance improvement.
    \item Our experiments show that although EagleEye is simple, it achieves the state-of-the-art pruning performance in comparisons with many more complex approaches. In the ResNet-50 experiments, EagleEye delivers 1.3\% to 3.8\% higher accuracy than compared algorithms. Even in the challenging task of pruning the compact model of MobileNet V1, EagleEye achieves the highest accuracy of 70.9\% with an overall 50\% operations (FLOPs) pruned. The results here are ImageNet top-1 classification accuracy.
\end{itemize}

\section{Related work}
\label{sec:related}
Pruning was mainly handled by hand-crafted heuristics in early time~\cite{fp}. So a pruned candidate network is obtained by human expertise and evaluated by training it to the converged accuracy, which can be very time consuming considering the large number of plausible sub-nets. In later chapters, we will show that the pruning candidate selection is problematic and selected pruned networks cannot necessarily deliver the highest accuracy after fine-tuning. Greedy strategy were introduced to save manual efforts~\cite{greedy} in more recent time. But it is easy for such strategy to fall into the local optimal caused by the greedy nature. For example, NetAdapt \cite{greedy} supposes the layer $l_t$ with the least accuracy drop, noted as $d_t$, is greedily pruned at step $t$. However, there may exist a better pruning strategy where $d'_{t} > d_{t}$, but $d'_{t} + d'_{t+1} < d_{t} + d_{t+1}$. Our method searches the pruning ratios for all layers together in one single step and therefore avoids this issue.

Some other works induce sparsity to weights in training phase for pruning purposes. For example, \cite{ssl} introduces group-LASSO to introduce sparsity of the kernels and ~\cite{networkslimming} regularizes the parameter in batch normalization layer. ~\cite{molchanov2019taylor} ranks the importance of filters based on Taylor expansion and trimmed off the low-ranked ones. The selection standards proposed in these methods are orthogonal to our proposed algorithm. More recently, versatile techniques were proposed to achieve automated and efficient pruning strategies such as reinforcement learning~\cite{amc}, generative adversarial learning mechanism~\cite{gal} and so on. But the introduced hyper-parameters add difficulty to repeat the experiments and the trail-and-error to get the auxiliary models work well can be time consuming.

The technique of adjusting BN was used to serve for non-pruning purposes in existing works. ~\cite{abn} adapts the BN statistics for target domain in domain adaptation tasks. The common point with our work is that we both notice the batch normalization requires an adjustment to adapt models in a new setting where either model or domain changes. But this useful technique has not been particularly used for model pruning purposes.
\section{Methodology}
\label{method}
\begin{figure}[h]
    \centering
       \includegraphics[width=0.9\linewidth]{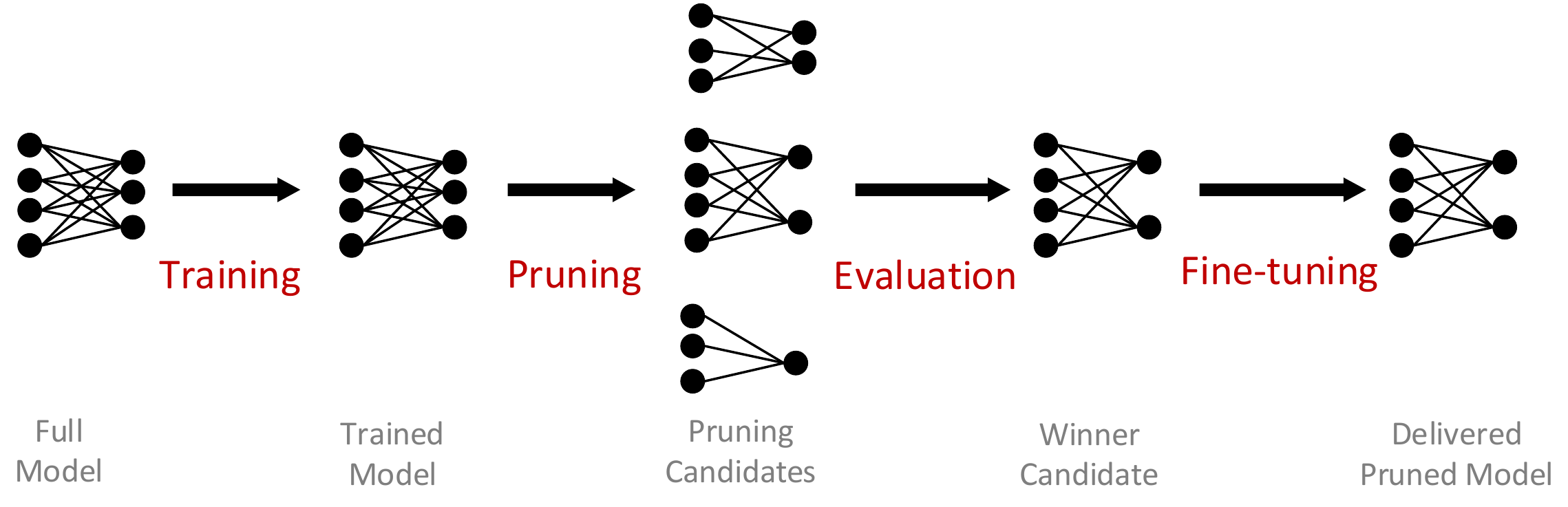}
       \caption{A typical pipeline for neural network training and pruning}
       \label{fig:pipeline}
\end{figure}

A typical neural network training and pruning pipeline is generalized and visualized in Figure~\ref{fig:pipeline}. Pruning is normally applied to a trained full-size network for redundancy removal purposes. An fine-tuning process is then followed up to gain accuracy back from losing parameters in the trimmed filters. In this work, we focus on structured filter pruning approaches, which can be generally formulated as
\begin{small}
\begin{equation}
\begin{aligned}
(r_1, r_2, ..., r_L)^* &= \mathop{\arg\min}_{r_1, r_2, ..., r_L}{\mathcal L(\mathcal A(r_1, r_2, ..., r_L;w))},\quad
 s.t.\ \mathcal C < constraints,
\end{aligned}
\label{eq:fp_eq}
\end{equation}\end{small}
where $\mathcal L$ is the loss function and $\mathcal A$ is the neural network model. $r_l$ is the pruning ratio applied to the $l^{th}$ layer. Given some constraints $\mathcal C$  such as targeted amount of parameters, operations, or execution latency, a combination of pruning ratios $(r_1, r_2, ..., r_L)$, which is referred as pruning strategy, is applied to the full-size model. All possible combinations of the pruning ratios form a searching space. To obtain a compact model with the highest accuracy, one should search through the search space by applying different pruning strategies to the model, fine-tuning each of the pruned model to converged and pick the best one. We consider the pruning task as finding the optimal pruning strategy, denoted as $(r_1, r_2, ..., r_L)^*$, that results in the highest converged accuracy of the pruned model.

Apart from handcraft designing, different searching methods have been applied in previous work to find the optimal pruning strategy, such as greedy algorithm~\cite{greedy,autoslim}, RL~\cite{amc}, and evlolutionary algorithm~\cite{metapruning}. All of the these methods are guided by the evaluation results of the pruning strategies.

\subsection{Motivation}
\label{sec:motivation}
\begin{figure}[htbp]
\centering
    \begin{minipage}[t]{0.48\textwidth}
        \centering
        \includegraphics[width=0.8\columnwidth]{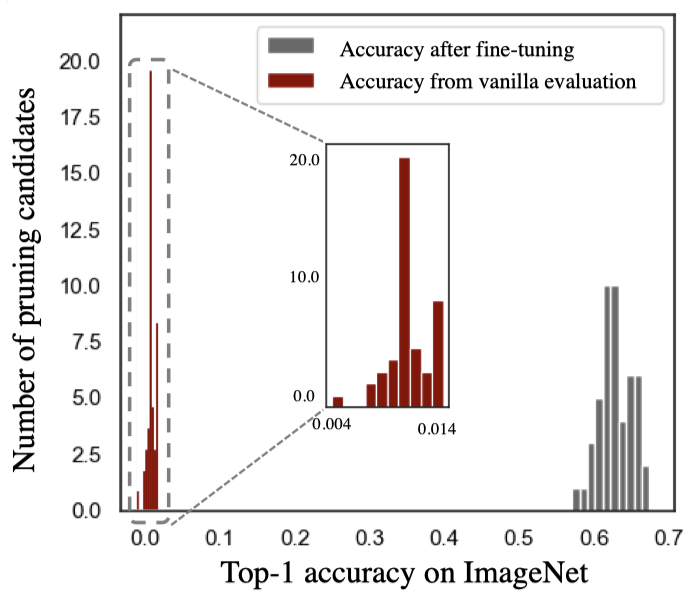}
        
    \end{minipage}
    \begin{minipage}[t]{0.48\textwidth}
        \centering
        \includegraphics[width=0.8\columnwidth]{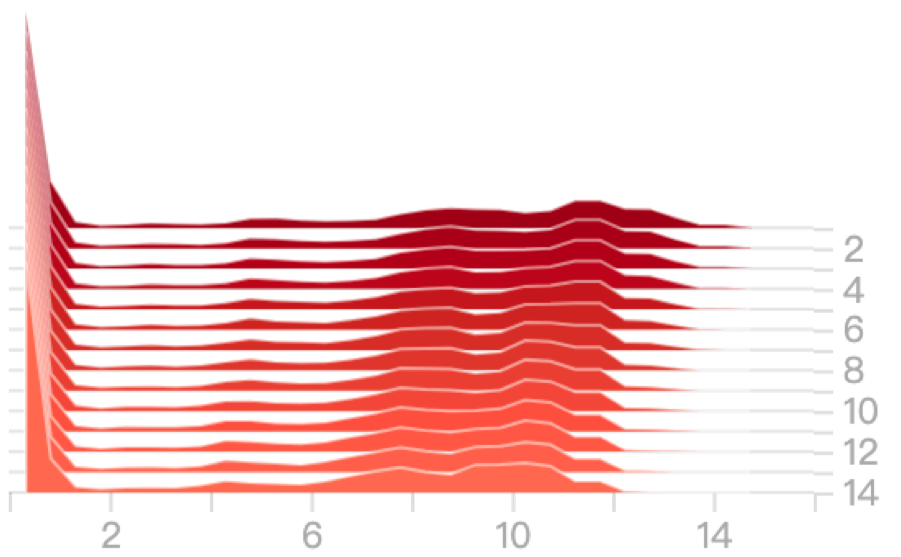}
    \end{minipage}
    \caption{\textbf{Left:}Histogram for accuracy collected from directly pruning MobileNet V1 and fine-tuning 15 epoches. \textbf{Right:}Evolution of the weight distribution of a pruned MobileNetV1~\cite{mv1} during fine-tuning on ImageNet~\cite{imagenet}. Where $X$ axis presents the magnitude of the $\mathcal{L}$1-norm of kernel, $Y$ axis presents the quantity, $Z$ axis presents the fine-tuning epochs.}
    \label{fig:acc_gap_param}
\end{figure}

In many published approaches~\cite{amc,fp,autocompress} in this domain, pruning candidates directly compare with each other in terms of evaluation accuracy. The sub-nets with higher evaluation accuracy are selected and expected to also deliver high accuracy after fine-tuning. However, such intention can not be necessarily achieved as we notice the sub-nets perform poorly if directly used to do inference. The inference results normally fall into a very low-range accuracy, which is illustrated in Figure~\ref{fig:acc_gap_param} left. An early attempt is to randomly generate pruning rates for MobileNet V1 and apply $\mathcal L1$-norm based pruning~\cite{fp} for 50 times. The dark red bars form the histogram of accuracy collected from directly doing inference with the pruned candidates in the same way that~\cite{amc,fp,autocompress} do before fine-tuning. Because our pruning rates are randomly generated in this early attempt, so the accuracy is very low and only for observation. The gray bars in Figure 4 shows the situation after fine-tuning these 50 pruned networks. We notice a huge difference in accuracy distribution between these two results. Therefore, there are two questions came up to our mind given above observation. The first question is why removal to filters, especially considered as “unimportant” filters, can cause such noticeable accuracy degradation although the pruning rates are random? The natural question to ask next is how strongly the low-range accuracy is positively correlated to the final converged accuracy. These two questions triggered our investigation into this commonly used evaluation process, which is called vanilla evaluation in this work.

Some initial investigations are done to tentatively address the above two questions. Figure~\ref{fig:acc_gap_param} right shows that it might not be the weights that mess up the accuracy at the evaluation stage as only a gentle shift in weight distribution is observed during fine-tuning, but the delivered inference accuracy is very different. On the other side, Figure~\ref{fig:cor_07pr} left shows that the low-range accuracy indeed presents poor correlation with the fine-tuned accuracy, which means that it can be misleading to use evaluated accuracy to guide the pruning candidates selection.

Interestingly, we found that it is the batch normalization layer that largely affects the evaluation. Without fine-tuning, pruning candidates have parameters that are a subset of those in the full-size model. So the layer-wise feature map data are also affected by the changed model dimensions. However, vanilla evaluation still uses Batch Normalization (BN) inherited from the full-size model. The outdated statistical values of BN layers eventually drag down the evaluation accuracy to a surprisingly low range and, more importantly, break the correlation between evaluation accuracy and the final converged accuracy of the pruning candidates in the strategy searching space. A brief training, also called fine-tuning, all pruning candidates and then compare them is a more accurate way to carry out the evaluation [20, 15]. However, it is very time-consuming to do the training-based evaluation for even single-epoch fine-tuning due to the large scale of the searching space. We give quantitative analysis later in this section to demonstrate this point.

Firstly, to quantitatively demonstrate the idea of vanilla evaluation and the problems that come with it, we symbolize the original BN~\cite{bn} as below:
\begin{small}\begin{equation}
    \begin{aligned}
    y &= \gamma \frac{x - \mu}{\sqrt{\sigma^2 + \epsilon}} + \beta, \\
    \end{aligned}
    \label{eq:bn_eq}
\end{equation}\end{small}Where $\beta$ and $\gamma$ are trainable scale and bias terms. $\epsilon$ is a term with small value to avoid zero division. For a mini-batch with size $N$, the statistical values of $\mu$ and $\sigma^2$ are calculated as below:\begin{small}\begin{equation}
    \begin{aligned}
    \mu_{\mathcal B} = E[x_{\mathcal B}] = \frac{1}{N}\sum_{i=1}^{N}x_i, \quad
    \sigma^2_{\mathcal B} = Var[x_{\mathcal B}] = \frac{1}{N-1}\sum_{i=1}^{N}(x_i - \mu_{\mathcal B})^2. \\
    \end{aligned}
    \label{eq:musig_train_eq}
\end{equation}\end{small}During training, $\mu$ and $\sigma^2$ are calculated with the moving mean and variance:
\begin{equation}
    \begin{aligned}
    \mu_{t} = m\mu_{t-1} + (1-m)\mu_{\mathcal B}, \quad
    \sigma^2_{t} = m\sigma^2_{t-1} + (1-m)\sigma_{\mathcal B}^2, \\
    \end{aligned}
    \label{eq:musig_test_eq}
\end{equation}
where $m$ is the momentum coefficient and subscript $t$ refers to the number of training iterations. In a typical training pipeline, if the total number of training iteration is $T$, $\mu_T$ and $\sigma_T^2$ are used in testing phase. These two items are called global BN statistics, where "global" refers to the full-size model.

\subsection{Adaptive Batch Normalization}
As briefly mentioned before, vanilla evaluation used in~\cite{amc,fp,autocompress} apply global BN statistics to pruned networks to fast evaluate their accuracy potential, which we think leads to the low-range accuracy results and unfair candidate selection. If the global BN statistics are out-dated to the sub-nets, we should re-calculate $\mu_T$ and $\sigma_T^2$ with adaptive values by conducting a few iterations of inference on part of the training set, which essentially adapts the BN statistical values to the pruned network connections.  Concretely, we freeze all the network parameters while resetting the moving average statistics. Then, we update the moving statistics by a few iterations of forward-propagation, using Equation \ref{eq:musig_test_eq}, but without backward propagation. We note the adaptive BN statistics as $\hat{\mu_T}$ and $\hat{\sigma_T^2}$.
\begin{figure}[]
  \centering
  \includegraphics[width=0.85\columnwidth]{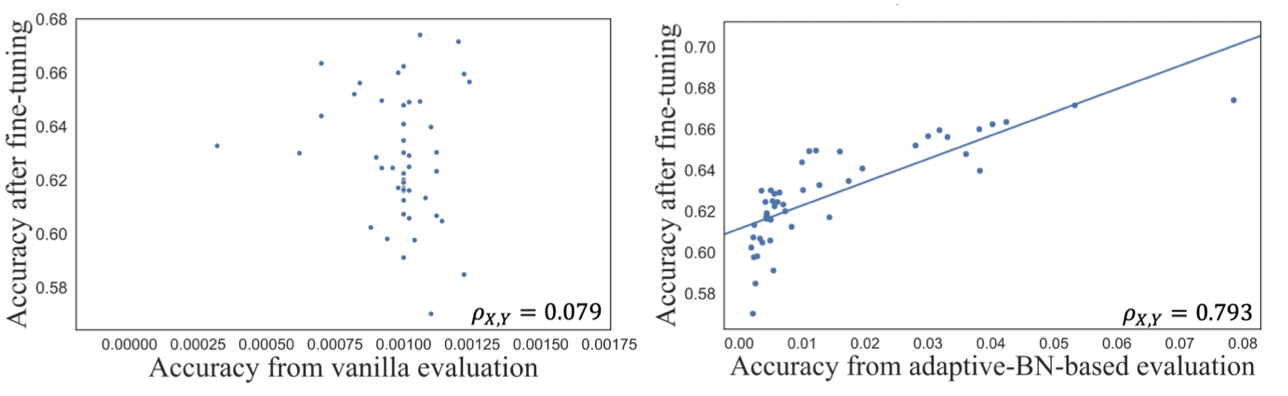}
\caption{Correlation between fine-tuning accuracy and inference accuracy gained from vanilla evaluation (left), adaptive-BN-based evaluation (right) based on MobileNet V1 experiments on ImageNet Top-1 classification results.}
\label{fig:cor_07pr}
\end{figure}

Figure~\ref{fig:cor_07pr} right illustrates that applying adaptive BN delivers evaluation accuracy that has a stronger correlation, compared to the vanilla evaluation Figure~\ref{fig:cor_07pr} left. 

\begin{figure}[t!]
    \centering
    \includegraphics[width=0.9\columnwidth]{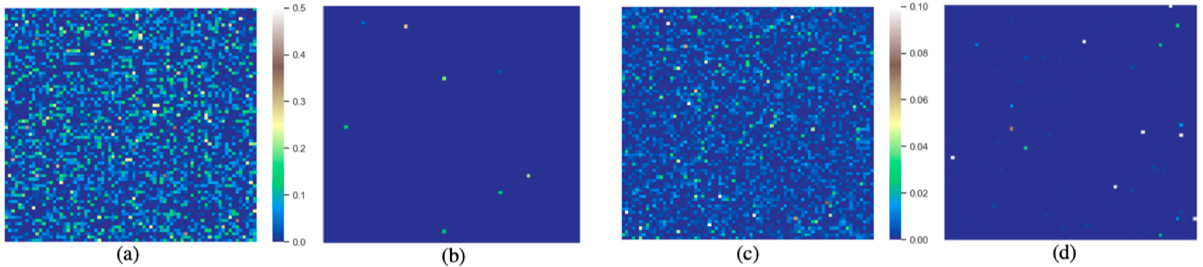}
    \caption{
        Visualization of distances of BN statistics in terms of the moving mean and variance. Each pixel refers to the distance of one BN statistics of a channel in MobileNetV1.
        \textbf{(a)} $\|{\mu_T} - \mu_{val}\|_2$, distance of moving mean between global BN and the “true” values.
        \textbf{(b)}  distance of moving mean between adaptive-BN and the “true” values $\|\hat{\mu_T} - \mu_{val}\|_2$.
        \textbf{(c)} $\left\|\sigma_{T}^{2}-\sigma_{v a l}^{2}\right\|_{2}$, distance of moving variance between global BN and the “true” values.
        \textbf{(d)} distance of moving variance between adaptive-BN and the “true” values$\left\|\sigma_{T}^{2}-\sigma_{v a l}^{2}\right\|_{2}$}
    \label{fig:bn_heatmap}
\end{figure}

As another evidence, we compare the distance of BN statistical values between “true” statistics. We consider $\mu$ and $\sigma^2$ sampled from the validation data as the “true” statistics, noted as $\mu_{val}$ and $\sigma_{val}^2$ , because they are the real statistical values in the testing phase. Specially, we are not obtaining insights from the validation data, which we think is unfair, but simply showing that our evaluation results are closer to the ground truth compared to the vanilla method. Concretely, we expect $\hat{\mu}_{T}$ and $\hat{\sigma}_{T}^{2}$ to be as close as possible to the “true” BN statistics values,$\mu_{val}$ and $\sigma_{v a l}^{2}$, so they could deliver close computational results. So we visualize the distance of BN statistical values gained from different evaluation methods (see Figure~\ref{fig:bn_heatmap}). Each pixel in the heatmaps represents a distance for a type of BN statistics, either $\mu_{val}$ or $\sigma_{val}^2$, between post-evaluation results and the “true” statistics sampled via one filter in MobileNet V1~\cite{mv1}. The visual observation shows that adaptive BN provides closer statistical values to the “true” values while global BN is way further. A possible explanation is that the global BN statistics are out-dated and not adapted to the pruned network connections. So they mess up the inference accuracy during evaluation for the pruned networks.

Noticeably, fine-tuning also relieves such problem of mismatched BN statistics because the training process itself re-calculates the BN statistical values in the forward pass and hence fixes the mismatch. However, BN statistics are not trainable values but sampling parameters only calculated in inference time. Our adaptive BN targets on this issue by conducting re-sampling in exactly the inference step, which achieves the same goal but with way less computational cost compared to fine-tuning. This is the main reason that we claim the application of adaptive BN in pruning evaluation is more efficient than the fine-tuning-based solution.

\subsection{Correlation Measurement}
\label{sec:corr_measurement}
As mentioned before, a “good” evaluation process in the pruning pipeline should present a strong positive correlation between the evaluated pruning candidates and their corresponding converged accuracy. Here, we compare two different evaluation methods, adaptive-BN-based and vanilla evaluation, and study their correlation with the fine-tuned accuracy. So we symbolize a vector of accuracy for all pruning candidates in the searching space (Figure~\ref{fig:fnnp_workflow}) separately using the above two evaluation methods as $X_1$ and $X_2$ correspondingly while fine-tuned accuracy is noted as $Y$.
We firstly use Pearson Correlation Coefficient~\cite{pearson}(PCC) $\rho_{X,Y}$, which is used to measure the linear correlation between two variables $X$ and $Y$, to measure the correlation between $\rho_{X_1,Y}$ and $\rho_{X_2,Y}$.

Since we particularly care about high-accuracy sub-nets in the ordered accuracy vectors, Spearman Correlation Coefficient (SCC)~\cite{spcnn18ICLR} $\phi_{X, Y}$ and Kendall rank Correlation Coefficient (KRCC)~\cite{Kendall1938ANM} $\tau_{X, Y}$ are adopted to measure the monotonic correlation.
We compare the correlation between $(X_1,Y)$ and $(X_2,Y)$ in above three metrics with different pruning rates. All cases present a stronger correlation for the adaptive-BN-based evaluation than the vanilla strategy.
See richer details about quantitative analysis in Section~\ref{sec:corr}.

\subsection{EagleEye pruning algorithm}
\begin{figure}[]
  \centering
  \includegraphics[width=0.9\columnwidth]{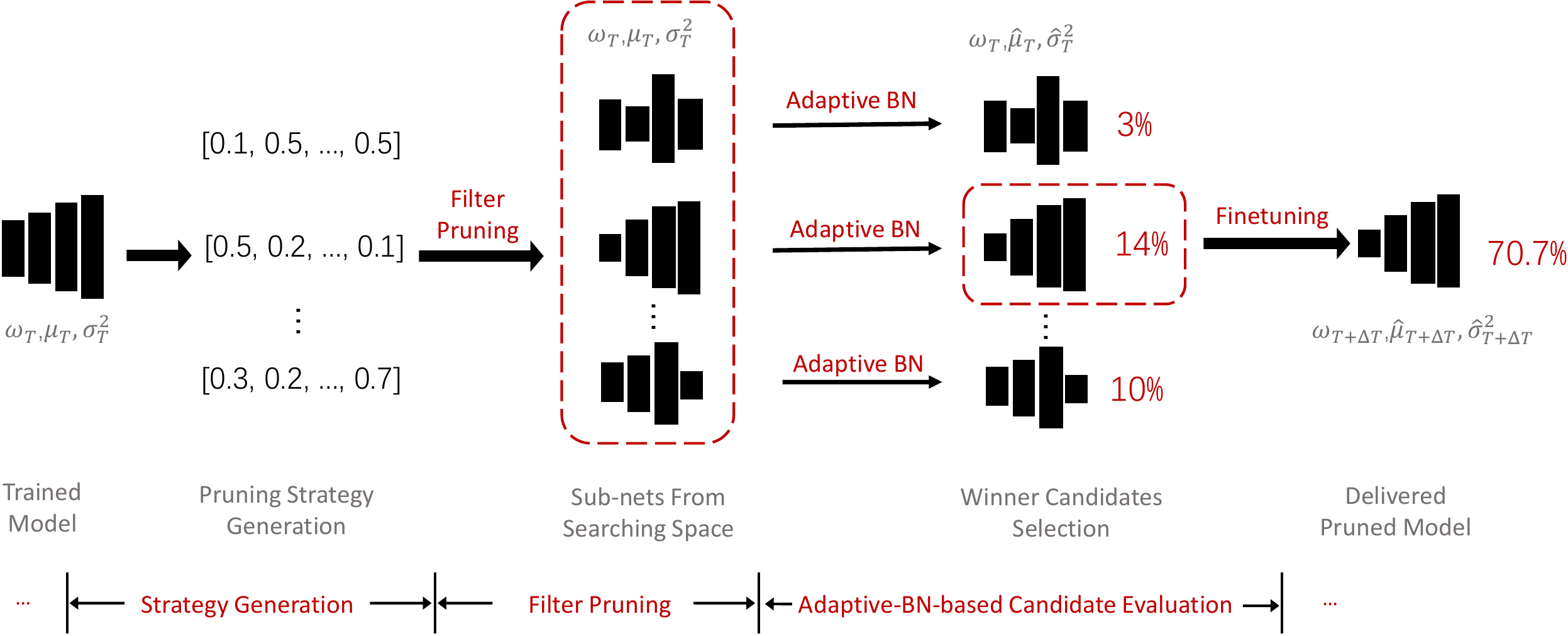}
\caption{Workflow of the EagleEye Pruning Algorithm}
\label{fig:fnnp_workflow}
\end{figure}

Based on the discussion about the accurate evaluation process in pruning, we now present the overall workﬂow of EagleEye in Figure~\ref{fig:fnnp_workflow}. Our pruning pipeline contains three parts, pruning strategy generation, filter pruning, and adaptive-BN-based evaluation.

\textbf{Strategy generation} outputs pruning strategies in the form of layer-wise pruning rate vectors like $(r_1, r_2, ..., r_L)$ for a $L$-layer model. The generation process follows pre-defined constraints such as inference latency, a global reduction of operations (FLOPs) or parameters and so on. Concretely, it randomly samples $L$ real numbers from a given range $[0, R]$ to form a pruning strategy, where $r_l$ denotes the pruning ratio for the $l^{th}$ layer. $R$ is the largest pruning ratio applied to a layer. This is essentially a Monte Carlo sampling process with a uniform distribution for all legitimate layer-wise pruning rates, i.e. removed number of filters over the number of total filters. Noticeably, other strategy generation methods can be used here, such as the evolutionary algorithm, reinforcement learning etc.,  we found that a simple random sampling is good enough for the entire pipeline to quickly yield pruning candidates with state-of-the-art accuracy. A possible reason for this can be that the adjustment to the BN statistics leads to a much more accurate prediction to the sub-nets' potential, so the efforts of generating candidates are allowed to be massively simplified. The low computation cost of this simple component also adds the advantage of fast speed to the entire algorithm.

\textbf{Filter pruning process} prunes the full-size trained model according to the generated pruning strategy from the previous module. Similar to a normal filter pruning method, the filters are firstly ranked according to their $\mathcal L1$-norm and the $r_l$ of the least important filters are trimmed off permanently. The sampled pruning candidates from the searching space are ready to be delivered to the next evaluation stage after this process.

\textbf{The adaptive-BN-based candidate evaluation module} provides a BN statistics adaptation and fast evaluation to the pruned candidates handed over from the previous module. Given a pruned network, it freezes all learnable parameters and traverses through a small amount of data in the training set to calculate the adaptive BN statistics $\hat{\mu}$ and $\hat{\sigma}^2$. In practice, we sampled 1/30 of the total training set for 100 iterations in our ImageNet experiments, which takes only 10-ish seconds in a single Nvidia 2080 Ti GPU. Next, this module evaluates the performance of the candidate networks on a small part of training set data, called ”sub-validation set”, and picks the top ones in the accuracy ranking as winner candidates. The correlation analysis presented in Section \ref{sec:corr} guarantees the effectiveness of this process. After a fine-tuning process, the winner candidates are finally delivered as outputs.
\section{Experiments}

\subsection{Quantitative analysis of correlation}
\label{sec:corr}

We use three commonly used correlation coefficient($\rho$,$\sigma$ and $\tau$) to quantitatively measure the relation between $X_1$, $X_2$ and $Y$ , which are defined in Section~\ref{sec:corr_measurement}.

\begin{figure}[]
    \centering
    \includegraphics[width=1\linewidth]{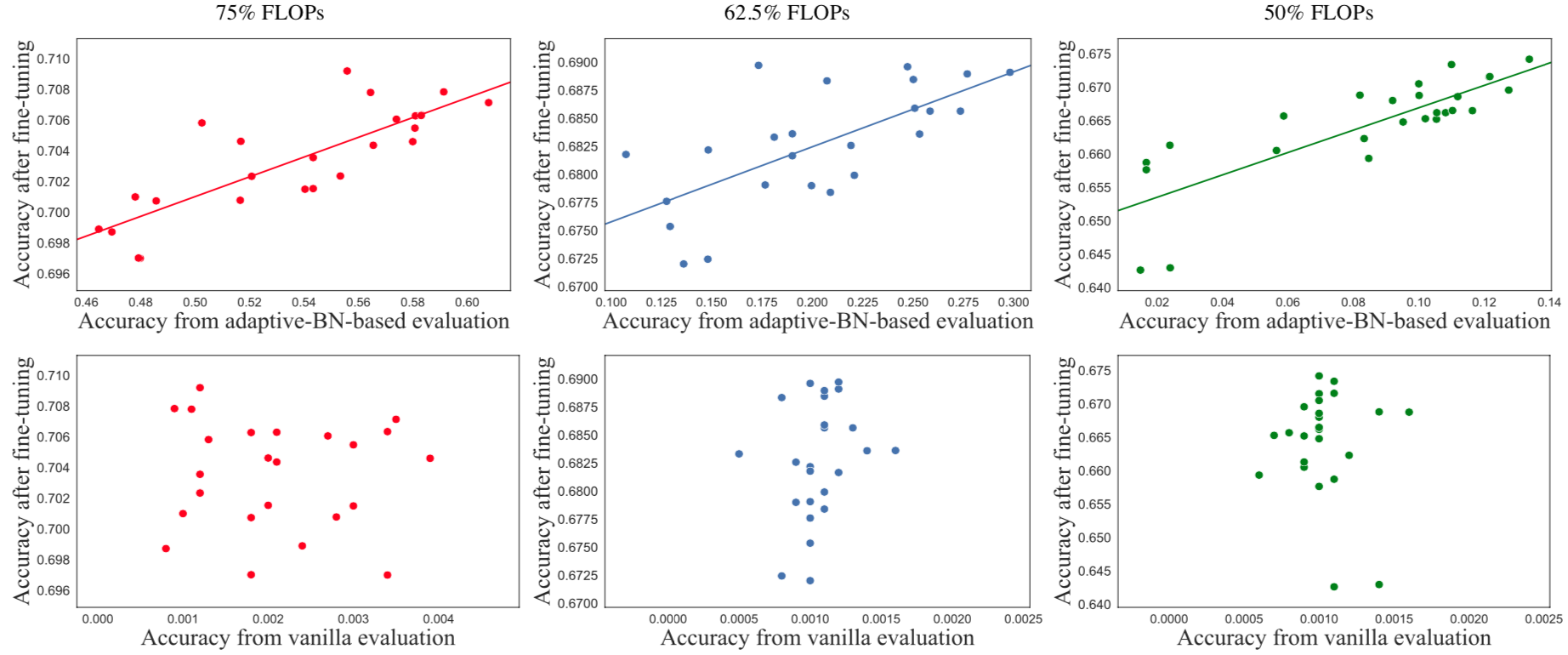}
    \caption{Vanilla vs. adaptive-BN evaluation: Correlation between evaluation and fine-tuning accuracy with different pruning ratios (MobileNet V1~\cite{mv1} on ImageNet~\cite{imagenet} classification Top-1 results)}
\label{fig:correlation}
\end{figure}

Firstly, as mentioned in Section~\ref{sec:motivation} the poor correlation, presented by Figure~\ref{fig:cor_07pr} sub-figure, is basically 10 times smaller than adaptive-BN-based results shown in Figure~\ref{fig:cor_07pr} right sub-figure. This matches with the visual observation that the adaptive-BN-based samples are more trendy while the vanilla strategy tends to give randomly distributed samples on the figure. This means the vanilla evaluation hardly present accurate prediction to the pruned networks about their fine-tuned accuracy.

Based on the above initial exploration, we extend the quantitative study to a larger scale applying three correlation coefficients to different pruning ratios as shown in Table~\ref{table:corr}. Firstly, the adaptive-BN-based evaluation delivers stronger correlation measured in all three coefficients compared to the vanilla evaluation. In average, $\rho$ is 0.67 higher, $\phi$ is 0.79 higher and $\tau$ is 0.46 higher. Noticeably, the correlation high in $\phi$ and $\tau$ means that the winner pruning candidates selected from the adaptive-based evaluation module are more likely to rank high in the fine-tuned accuracy ranking as $\phi$ emphasizes the monotonic correlation.

Especially, the third to fifth rows of Table~\ref{table:corr} shows the correlation metrics with different pruning rates (for instance, 75\% FLOPs also means 25\% pruning rate to operations). The corresponding results are also visualized in Figure~\ref{fig:correlation}. The second row in Table~\ref{table:corr} means the pruning rate follows a layer-wise Monte Carlo sampling with a uniform distribution among the legitimate pruning rate options. All the above tables and figures prove that the adaptive-BN-based evaluation shows stronger correlation, and hence a more robust prediction, between the evaluated and fine-tuned accuracy for the pruning candidates.

\begin{table}[t]
\renewcommand\arraystretch{1.2}
\centering
\caption{Correlation analysis quantified by Pearson Correlation Coefficient $\rho_{X,Y}$, Spearman Correlation Coefficient $\phi_{X,Y}$ , and Kendall rank Correlation Coefficient $\tau_{X,Y}$.}
    \begin{tabular}{c|c|c||c|c||c|c}
    \hline
    FLOPs constraints & $\rho_{X_1,Y}$ & $\rho_{X_2,Y}$ & $\phi_{X_1,Y}$ & $\phi_{X_2,Y}$ & $\tau_{X_1,Y}$ & $\tau_{X_2,Y}$ \\ \hline
    Not Fixed         & 0.793          & 0.079          & 0.850            & 0.025            & 0.679          & 0.063          \\ \hline
    75\% FLOPs        & 0.819          & -0.038         & 0.829            & -0.030           & 0.656          & -0.003         \\
    62.5\% FLOPs      & 0.683          & 0.250          & 0.644            & 0.395            & 0.458          & 0.267          \\
    50\% FLOPs        & 0.813          & 0.105          & 0.803            & 0.127            & 0.639          & 0.122          \\ \hline
    \end{tabular}
    \label{table:corr}
\end{table}
\subsection{Generality of the adaptive-BN-based evaluation method}
The proposed adaptive-BN-based evaluation method is general enough to plug-in and improves some existing methods. As an example, we apply it to AMC~\cite{amc}, which is an automatic method based on Reinforcement Learning  mechanism.

AMC~\cite{amc} trains an RL-agent to decide the pruning ratio for each layer. At each training step, the agent tries applying different pruning ratios (pruning strategy) to the full-size model as an action. Then it directly evaluates the accuracy without fine-tuning, which is noted as vanilla evaluation in our paper, and takes this validation accuracy as the reward. As the RL-agent is trained with the reward based on the vanilla evaluation, which is proved to have a poor correlation to the converged accuracy of pruned networks. So we replace the vanilla evaluation process with our proposed adaptive-BN-based evaluation. Concretely, after pruning out filters at each step, we freeze all learnable parameters and do inference on the training set to fix the BN statistics and evaluate the accuracy of the model on the sub-validation set. We feed this accuracy as a reward to train the RL-agent in place of the accuracy of vanilla evaluation. The experiment about MobileNetV1~\cite{mv1} on ImageNet~\cite{imagenet} classification accuracy is improved from 70.5\% (reported in AMC~\cite{amc}) to 70.7\%. It shows that the RL-agent can find a better pruning strategy with the help of our adaptive-BN-based evaluation module.

Another example is the “short-term fine-tune” block in~\cite{greedy}, which also can be handily replaced by our adaptiveBN-based module for a faster pruning strategy selection. On the other side, our pipeline can also be upgraded by existing methods such as the evolutionary algorithm used in~\cite{metapruning} to improve the basic Monte Carlo sampling strategy. The above experiments and discussion demonstrate the generality of our adaptive-BN-based evaluation module, but can not be analyzed in more detail due to the limited length of this paper.

\subsection{Efficiency of our proposed method}
\begin{table}[]
\renewcommand\arraystretch{1.2}
        \centering
        \caption{Comparison of computation costs of various pruning methods in the task where all pruning methods are executed to find the best pruning strategy from 1000 potential strategies (candidates).}
            \resizebox{\textwidth}{!}{\begin{tabular}{c|c|c|c}
                \hline
                \textbf{Method} & \textbf{Evaluation Method} & \textbf{Candidate Selection} & \textbf{GPU Hours}\\
                \hline
                ThiNet~\cite{thinet} & finetuning & 1000$\times$10 finetune epochs & $\sim$ 8000\\
                NetAdapt~\cite{greedy} & finetuning & $10^4$ training iterations & 864\\
                Filter Pruning~\cite{fp} & vanilla & 1000$\times$25 finetune epochs & $\sim$ 20000\\
                \hline
                AMC~\cite{greedy} & vanilla & Training an RL agent & -\\
                Meta-Pruning~\cite{metapruning} & PruningNet & Training an auxiliary network & -\\
                \hline
                \textbf{EagleEye} & adaptive-BN & $<$1000$\times$100 inference iterations& 25\\
                \hline
            \end{tabular}}
        \label{tab:eva_comp}
\end{table}
Our proposed pruning evaluation based on adaptive BN turn the prediction of sub-net accuracy into a very fast and reliable process, so EagleEye is much less time-consuming to complete the entire pruning pipeline than other heavy evaluation based algorithms. In this part, we compare the execution cost for various state-of-the-art algorithms to demonstrate the efficiency of our method.

Table~\ref{tab:eva_comp} compares the computational costs of picking the best pruning strategy among 1000 potential pruning candidates. As ThiNet~\cite{thinet} and Filter Pruning~\cite{fp} require manually assigning layer-wise pruning ratio, The final GPU hours are the estimation of completing the pruning pipeline for 1000 random strategies. In practice, the real computation cost highly depends on the expert's heuristic practice of trial-and-error. The computation time for AMC~\cite{amc} and Meta-pruning can be long because training either an RL network or an auxiliary network itself is time-consuming and tricky. Among all compared methods, EagleEye is the most efficient method as each evaluation takes no more than 100 iterations, which takes 10 to 20 seconds in a single Nvidia 2080 Ti GPU. So the total candidate selection is simply an evaluation comparison process, which also can be done in negligible time.

\subsection{Effectiveness of our proposed method}
To demonstrate the effectiveness of EagleEye, we compare it with several state-of-the-art pruning methods on MobileNetV1 and ResNet-50~\cite{resnet} models tested on the small dataset of CIFAR-10~\cite{cifar} and the large dataset of ImageNet.

\begin{table}[]
\caption{Pruning results of ResNet-56 (left) and MobileNetV1 (right) on CIFAR-10} 
\centering
    \begin{minipage}[t]{0.48\textwidth}
        \centering
        \begin{tabular}{ccc}
             \hline
             Method & FLOPs & Top1-Acc \\
             \hline
             \hline
             ResNet-56 & 125.49M & 93.26\% \\
             FP~\cite{fp} & 90.90M & 93.06\% \\
             RFP~\cite{Ayinde2018BuildingEC} & 90.70M & 93.12\%\\
             NISP~\cite{Yu2017NISPPN} & 81.00M & 93.01\% \\
             GAL~\cite{Lin2019TowardsOS} & 78.30M & 92.98\% \\
             HRank~\cite{Lin2020HRankFP} & 88.72M & 93.52\% \\
             \textbf{EagleEye} & 62.23M & \textbf{94.66}\% \\ \hline \hline
        \end{tabular}
    \end{minipage}
    \begin{minipage}[t]{0.48\textwidth}
        \centering
        \begin{tabular}{ccc}
            \hline
            Method & FLOPs & Top1-Acc \\
            \hline
            \hline
            0.75 $\times$ MobileNetV1 & \multirow{3}{*}{26.5M} & 88.07\%\\
            FP(our-implement)~\cite{fp} & & 91.58 \% \\
            \textbf{EagleEye} & & \textbf{91.89}\% \\ \hline
            0.5 $\times$ MobileNetV1  & \multirow{3}{*}{12.1M} & 87.51\%\\
            FP(our-implement)~\cite{fp} & & 90.4\% \\
            \textbf{EagleEye} && \textbf{91.44}\% \\ \hline
            0.25 $\times$ MobileNetV1 & \multirow{3}{*}{3.3M} & 84.59\%\\
            FP(our-implement)~\cite{fp} && 85.81\% \\
            \textbf{EagleEye} && \textbf{88.01}\% \\ \hline
            \hline
        \end{tabular}
    \end{minipage}
    \label{tab:sota_cifar}
\end{table}

\begin{table}[h]
\small
    \centering
    \caption{Comparisions of ResNet-50 and other pruning methods on ImageNet}
        \begin{tabular}{ccccc}
            \hline
            FLOPs after pruning & Method & FLOPs & Top1-Acc & Top5-Acc\\
            \hline
            \hline
            \multirow{4}{*}{3G}
               & ThiNet-70~\cite{metapruning}   & 2.9G & 75.8\% & 90.67\% \\
               & AutoSlim~\cite{autoslim} & 3.0G & 76.0\% & - \\
               & Meta-Pruning~\cite{metapruning} & 3.0G & 76.2\% & - \\
               & \textbf{EagleEye}        & 3.0G & \textbf{77.1\%} & \textbf{93.37\%} \\
            \hline
            \multirow{11}{*}{2G}
               & 0.75 $\times$ ResNet-50~\cite{resnet} & 2.3G & 74.8\% & -\\
               & Thinet-50~\cite{thinet}   & 2.1G & 74.7\% & 90.02\% \\
               & AutoSlim~\cite{autoslim} & 2.0G & 75.6\% & -\\
              & CP~\cite{cp} & 2.0G & 73.3\% & 90.8\% \\
              & FPGM~\cite{fpgm} & 2.31G & 75.59\% & 92.63\% \\
              & SFP~\cite{sfp} & 2.32G & 74.61\% & 92.06\%\\
              & GBN~\cite{gbn} & 1.79G & 75.18\% & 92.41\% \\
               & GDP~\cite{gdp} & 2.24G & 72.61\% & 91.05\% \\
               & DCP~\cite{dcp} & 1.77G & 74.95\% & 92.32\% \\
               & Meta-Pruning~\cite{metapruning} & 2.0G & 75.4\% & - \\
               & \textbf{EagleEye} & 2.0G & \textbf{76.4\%} & \textbf{92.89\%} \\
            \hline
            \multirow{5}{*}{1G}
               & 0.5 $\times$ ResNet-50~\cite{resnet} & 1.1G & 72.0\% & -\\
               & ThiNet-30~\cite{thinet}   & 1.2G & 72.1\% & 88.30\% \\
               & AutoSlim~\cite{autoslim} & 1.0G & 74.0\% & -\\
               & Meta-Pruning~\cite{metapruning} & 1.0G & 73.4\% & -\\
               & \textbf{EagleEye}        & 1.0G & \textbf{74.2}\% & \textbf{91.77\%} \\
            \hline
        \end{tabular}
    \label{tab:sota_res}
\end{table}

\textbf{ResNet}
Table~\ref{tab:sota_cifar} left shows EagleEye outperforms all compared methods in terms of Top-1 accuracy on CIFAR-10 dataset. To further prove the robustness of our method, we compare the top-1 accuracy of ResNet-50 on ImageNet under different FLOPs constraints. For each FLOPs constraint (3G, 2G, and 1G), 1000 pruning strategies are generated. Then the adaptive-BN-based evaluation method is applied to each candidate. We just fine-tune the top-2 candidates and return the best as delivered pruned model. It is shown that EagleEye achieves the best results among the compared approaches listed in Table~\ref{tab:sota_res}.

ThiNet~\cite{thinet} prunes the channels uniformly for each layer other than finding an optimal pruning strategy, which hurts the performance significantly. Meta-Pruning~\cite{metapruning} trains an auxiliary network called ``PruningNet" to predict the weights of the pruned model. But the adopted vanilla evaluation may mislead the searching of the pruning strategies. As shown in Table~\ref{tab:sota_res}, our proposed algorithm outperform all compared methods given different pruned network targets.

\textbf{MobileNet} 
We conduct experiments of the compact model of MobileNetV1 and compare the pruning results with Filter Pruning~\cite{fp} and the directly-scaled models. Please refer to supplementary material for more details about FP implementation and training methods to get the accuracy for the directly-scaled models. Table~\ref{tab:sota_cifar} right shows that EagleEye gets the best results in all cases. 

Pruning MobileNetV1 for ImageNet is more challenging as it is already a very compact model. We compare the top-1 ImageNet classification accuracy under the same FLOPs constraint (about 280M FLOPs) and the results are shown in Table~\ref{tab:sota_mob}. 1500 pruning strategies are generated with this FLOPs constraint. Then adaptive-BN-based evaluation is applied to each candidate. After fine-tuning the top-2 candidates, the pruning candidate that returns the highest accuracy is selected as the final output.

AMC~\cite{amc} trains their pruning strategy decision agent based on the pruned model without fine-tuning, which may lead to a problematic selection on the candidates. NetAdapt~\cite{greedy} searches for the pruning strategy based on a greedy algorithm, which may drop into a local optimum as analysed in Section \ref{sec:related}. It is shown that EagleEye achieves the best performance among all studied methods again in this task (see Table~\ref{tab:sota_mob}).

\begin{table}[t]
\renewcommand\arraystretch{1.2}
    \centering
    \caption{Comparisions of MobileNetV1 and other pruning methods on ImageNet}
        \begin{tabular}{ccccc}
            \hline
            Method & FLOPs & Top1-Acc & Top5-Acc\\
            \hline
            \hline
            0.75 $\times$ MobileNetV1~\cite{mv1} & 325M & 68.4\% & -\\
            AMC~\cite{amc} & 285M & 70.5\% & - \\
            NetAdapt~\cite{greedy} & 284M & 69.1\% & - \\
            Meta-Pruning~\cite{metapruning}   & 281M & 70.6\% & - \\
            \textbf{EagleEye}        & 284M & \textbf{70.9}\% & \textbf{89.62}\%\\
            \hline
        \end{tabular}
    \label{tab:sota_mob}
\end{table}

\section{Discussion and Conclusions}

We presented EagleEye pruning algorithm, in which a fast and accurate evaluation process based on adaptive batch normalization is proposed. Our experiments show the efficiency and effectiveness of our proposed method by delivering higher accuracy than the studied methods in the pruning experiments on ImageNet dataset. An interesting work is to further explore the generality of the adaptive-BN-based module by integrating it into many other existing methods and observe the potential improvement. Another experiment that is worth a try is to replace the random generation of pruning strategy with more advanced methods such as evolutionary algorithms and so on.

\section*{Acknowledgements}
Jiang Su is the corresponding author of this work. This work was supported in part by the National Natural Science Foundation of China (NSFC) under Grant No.U1811463.
%
%
\bibliographystyle{splncs04}
\bibliography{egbib}
\end{document}